# Diversified Multi-prototype Representation for Semi-supervised Segmentation

**Jizong Peng**[*]
ETS Montreal
jizong.peng.1@etsmtl.net

**Christian Desrosiers**
ETS Montreal
christian.desrosiers@etsmtl.ca

**Marco Pedersoli**
ETS Montreal
marco.pedersoli@etsmtl.ca

## Abstract

This work considers semi-supervised segmentation as a dense prediction problem based on prototype vector correlation, and proposes a simple way to represent each segmentation class with multiple prototypes. To avoid degenerate solutions, two regularization strategies are applied on unlabeled images. The first one leverages mutual information maximization to ensure that all prototype vectors are considered by the network. The second explicitly enforces prototypes to be orthogonal by minimizing their cosine distance. Experimental results on two benchmark medical segmentation datasets reveal our method's effectiveness in improving segmentation performance when few annotated images are available.

## 1  Introduction

Supervised learning approaches based on convolutional neural networks (CNNs) have achieved outstanding performance in a wide range of segmentation tasks. However, these approaches often require a large amount of labeled images, difficult to obtain for medical imaging applications. Semi-supervised learning (SSL) is commonly used to reduce the need for fully-annotated data by exploiting unlabeled ones. In recent years, a broad range of semi-supervised methods have been proposed for medical image segmentation, including approaches based on transformation consistency [1, 2, 3], co-training of multiple models [4, 5, 6] and adversarial learning [7, 8]. While different from image classification, segmentation can be viewed as a dense prediction problem where classification is performed at the pixel level. Based on this idea, recent works have used metric learning [9, 10] to learn a robust representation for this task. These methods were shown to be effective for cardiac [11, 10, 12] and prostate [13, 10] segmentation. The core principle of these methods is to minimize a distance metric for similar pixels (*positive* pairs) while minimizing it for dissimilar ones (*negative* pairs). These pairs can be defined by image region [12, 10, 11] or segmentation boundary [13].

In this work, we consider segmentation as the problem of learning a fixed number of prototype vectors, modeled as $1 \times 1$ convolution kernels in the last convolution layer. During training, prototype vectors of a given class are updated so that they have high correlation to feature vectors located in regions corresponding to that class. We argue that widely-used segmentation networks such as U-Net [14] are designed to have only one prototype vector per class, and that this may be insufficient to capture the full variability of a class. Instead, we assign multiple prototype vectors to the same class by increasing the number of output classes. This simple strategy leads to an over-segmented prediction that can then be converted to a normal segmentation by learning the right prototype-to-class correspondence. To avoid trivial solution for the over-segmented prediction, we introduce two regularization losses based on mutual information maximization and prototype vector orthogonality. The former one ensures

---

that prototypes are used in a fair manner, while the other imposes them to be uncorrelated. These losses increase the usefulness and diversity of prototypes, which leads to a more robust prediction. We note that this differs from the Prototypical Network approach [15] for few-shot learning, where prototype vectors are obtained by averaging the representations of examples in the same class, similar to K-Means [16]. We validate our proposed method on two benchmark medical image datasets, evaluating the binary and multi-class segmentation of different organs. Results show our method to outperform various baselines and recent approaches for semi-supervised segmentation, when few labeled images are provided.

## 2 Method

In semi-supervised segmentation, we are given a small set labeled examples $\mathcal{D}_\ell = \{(\mathbf{x}_\ell, \mathbf{y}_\ell)\}$, each composed of an image $\mathbf{x} \in \mathbb{R}^{|\Omega|}$ and segmentation ground-truth $\mathbf{y}_\ell \in \{0,1\}^{|\Omega| \times |C|}$, and a larger set of unlabeled images $\mathcal{D}_u = \{\mathbf{x}_u\}$. Here, $\Omega = \{1, \ldots, W \times H\}$ is the set of image pixels indexes and $C$ the number of segmentation classes. We seek a segmentation function $f$ with parameters $\theta$ that takes as input an image $\mathbf{x}$ and returns for each pixel $i$ and class label $k$ a probability $f_{ik}(\mathbf{x})$. A CNN-based network is typically used for $f$. Such architecture decomposes the segmentation function as $f = (g \circ \phi)$, where $\phi(\mathbf{x}) \in \mathbb{R}^{|\Omega| \times N}$ is a feature map obtained obtained from CNN layers and $g$ is a function projecting the features to class probabilities, typically implemented using a $1 \times 1$ convolution followed by a softmax. The final output can thus be expressed as $f_{ik}(\mathbf{x}) = \mathrm{softmax}\big(\mathbf{w}_k^\top \phi_i(\mathbf{x})\big)$ where $\mathbf{w}_k$ is the $k$-th column of a kernel matrix $\mathbf{W} \in \mathbb{R}^{N \times C}$. As in most few shot learning approaches [15], we can consider $\mathbf{w}_k$ as a prototype vector for class $k$. The probability of pixel $i$ to be mapped to class $k$ is then proportional to the correlation between its feature vector $\phi_i(\mathbf{x})$ and prototype $\mathbf{w}_k$.

A problem with this simple model is its strong assumption that the variability of a class can be fully captured with a single prototype. We argue that this limited class representation is sub-optimal, especially when few annotated images are available for training. A straightforward way to alleviate this problem, which has not been well explored in the medical imaging community, is to represent each class with multiple prototype vectors. Toward this goal, we augment the number of output classes from $C$ to $C'$, with $C' > C$, which leads to an over-segmentation of the input image. Denote as $f'(\mathbf{x}) \in [0,1]^{|\Omega| \times C'}$ this new output, one can then recover the final segmentation by finding a mapping $\mathcal{M} : \{1, \ldots, C'\} \to \{1, \ldots, C\}$ from the over-segmentation classes to the real ones. While this mapping can be learned, in our method, we assume that each class is represented by the same number of prototypes and define $\mathcal{M}$ as a fixed $C' \times C$ matrix $\mathbf{M}$ such that $m_{jk} = 1$ if over-segmentation class $j$ is mapped to real class $k$, else $m_{jk} = 0$. The probability for pixel $i$ and class $k$ is then obtained as $f_{ik}(\mathbf{x}) = \sum_{j=1}^{C'} m_{jk} f'_{ij}(\mathbf{x})$. We leverage annotated images with a supervised loss measuring the cross-entropy between the ground-truth labels and the real class probabilities:

$$\mathcal{L}_{\mathrm{sup}}(\theta; \mathcal{D}_\ell) \,=\, \frac{1}{|\mathcal{D}_\ell|} \sum_{(\mathbf{x},\mathbf{y}) \in \mathcal{D}_\ell} \ell_{\mathrm{CE}}\big(\mathbf{y}, f(\mathbf{x})\big), \ \text{where} \ \ell_{\mathrm{CE}}(\mathbf{y}, \mathbf{p}) \,=\, -\frac{1}{|\Omega|} \sum_{i=1}^{|\Omega|} \sum_{k=1}^{C} y_{ik} \log(p_{ik}). \quad (1)$$

However, optimizing this supervised loss does not guarantee that the prototypes of over-segmentation classes model distinct and useful properties of their underlying real class. To ensure this, we add two regularization losses on unlabeled examples, based on mutual information maximization and prototype orthogonality.

**Mutual Information based regularization** Since the number of over-segmentation prototypes exceeds the number of real classes, it can happen that some of these prototypes are ignored during optimization and, thus, have a low output probability. To ensure that all prototypes are used and that their marginal distribution is balanced, we maximize the mutual information $\mathcal{I}(X; Y')$ between a random variable $X$ modeling an image and a random variable $Y'$ encoding its over-segmentation class label. Using the fact that $\mathcal{I}(X; Y') = \mathcal{H}(Y') - \mathcal{H}(Y'|X)$, where $\mathcal{H}(\mathbf{p}) = -\frac{1}{|\Omega|} \sum_i \sum_k p_k \log(p_{ik})$ is the Shannon entropy (averaged over pixels), we define the first regularization loss as

$$\mathcal{L}_{\mathrm{MI}}(\theta; \mathcal{D}_u) \,=\, -\mathcal{H}\bigg(\frac{1}{|\mathcal{D}_u|} \sum_{\mathbf{x} \in \mathcal{D}_u} f'(\mathbf{x})\bigg) + \frac{1}{|\mathcal{D}_u|} \sum_{\mathbf{x} \in \mathcal{D}_u} \mathcal{H}\big(f'(\mathbf{x})\big). \quad (2)$$

Table 1: Mean 3D DSC over three independent runs of tested methods.

Figure 1: Visual results. From left to right: ground truth, baseline, ours, and over-segmented prediction.

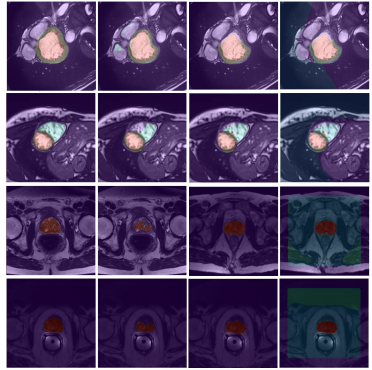

| Method | MP | MI | Orth | PROMISE12 | | ACDC | | |
|---|---|---|---|---|---|---|---|---|
| | | | | 3 scans | 5 scans | 1 scan | 2 scans | 4 scans |
| Full Sup. | | | | 88.06 | | 89.70 | | |
| Baseline | | | | 52.33 | 65.17 | 60.77 | 74.50 | 78.91 |
| Variant-1 | ✓ | | | 53.20 | 70.95 | 65.25 | 75.95 | 80.56 |
| Variant-2 | | ✓ | | 71.52 | 79.50 | 69.91 | 80.75 | 81.44 |
| Variant-3 | | | ✓ | 52.55 | 66.55 | 64.53 | 74.15 | 80.74 |
| Variant-4 | ✓ | ✓ | | **73.88** | 78.30 | 72.00 | 80.68 | 83.01 |
| Variant-5 | ✓ | | ✓ | 55.01 | 70.95 | 66.58 | 77.17 | 78.97 |
| Ours | ✓ | ✓ | ✓ | 73.38 | **79.69** | **72.44** | **82.11** | **83.09** |
| Entropy Min. [18] | | | | 53.37 | 69.74 | 67.53 | 76.08 | 80.59 |
| Pseudo Label [19] | | | | 53.50 | 67.40 | 65.74 | 75.33 | 79.59 |
| Mean Teacher [1] | | | | 65.73 | 78.79 | 81.61 | 84.00 | **84.72** |
| MT + Ours | ✓ | ✓ | ✓ | **75.30** | **80.26** | **82.10** | **85.70** | 84.53 |

As described in [17], minimizing the first term of $\mathcal{L}_{\text{MI}}$ increases the entropy of the marginal distribution, defined as the average of individual predictions, making it more uniform. On the other hand, the second term encourages the network to have confident predictions for the over-segmentation classes.

**Orthogonality regularization** To learn prototype vectors representing distinct characteristics, we also impose them to be uncorrelated. This is achieved with our prototype orthogonality loss:

$$\mathcal{L}_{\text{orth}}(\theta) = \|\mathbf{W}\mathbf{W}^\top - \mathbf{I}_{C'}\|_F^2 \qquad (3)$$

where $\mathbf{I}$ is the $C' \times C'$ identity matrix.

**Complete objective** The total loss, including the supervised loss and the two unsupervised regularization losses, is finally defined as

$$\mathcal{L}_{\text{total}} = \mathcal{L}_{\text{sup}}(\theta; \mathcal{D}_l) + \lambda_1 \mathcal{L}_{\text{MI}}(\theta; \mathcal{D}_u) + \lambda_2 \mathcal{L}_{\text{orth}}(\theta) \qquad (4)$$

where $\lambda_1, \lambda_2 \geq 0$ are coefficients balancing the different loss terms.

## 3 Experiments

**Dataset and evaluation metric** We evaluate our proposed method and its variants on two benchmark datasets for medical image segmentation: PROMISE12 and ACDC. PROMISE12 consists of 50 prostate MRI scans from different patients, which are split randomly in a training, validation and test sets containing 40, 3, and 7 scans, respectively. ACDC contains 200 cardiac cine-MRI scans and corresponding ground-truth segmentation masks for three classes: left ventricle (LV), right ventricle (RV) and myocardium of LV. We randomly selected 174, 9, and 17 scans of this dataset for training, validation and testing. The same prepossessing protocol and data augmentation were used as in [20]. For semi-supervised learning, we randomly chose a few scan from their training set as labeled set and considered others as unlabeled. We report the mean 3D Dice score (DSC) on the test set over 3 runs.

**Implementation details** We choose the 2D UNet [14] as our main architecture and optimize it using stochastic gradient descent (SGD) with the RAdam optimizer [21]. For all experiments, we used the public code base [b] from [20], keeping the same optimization strategy. Using a grid search on the validation set, we selected the following hyper-parameters: 3 prototype vectors per class, $\lambda_1 = 0.01$ and $\lambda_2 = 0.5$.

**Results** The upper part of Table 1 reports the performances of baselines and ablation variants of our method. *Full Sup.* is an upper bound where all available training data are used as labeled examples. Likewise, *Baseline* serves as our lower bound where only a few labeled data are used (see Table) and unlabeled images are ignored. We tested six variants of our method using multiple prototypes or a single one (MP), the mutual information loss (MI), and the prototype orthogonality loss (Orth). As can be seen, using MP, MI, and Orth together leads to the greatest improvement over the Baseline in all but one cases, boosting DSC by 21.1% for PROSTATE12 with 3 labeled scans, and by 11.7% for ACDC with one labeled scan. Visual results are given in Fig. 1, showing examples of the predicted over-segmentation and its benefit on the final segmentation. We then compare our method to three popular semi-supervised segmentation approaches, [18], pseudo-label [19], and Mean Teacher (MT) [1]. Results show our method to outperform Entropy minimization and pseudo-label for both datasets, while MT is better for ACDC. Based on this observation, we boost our method by adding a temporal ensembling strategy on top as in Mean Teacher. Results at the bottom of Tab. 1 show a significant improvement over standard Mean Teacher when using very few labeled images.

---
[b] https://github.com/jizongFox/MI-based-Regularized-Semi-supervised-Segmentation

## 4 Social impact

The proposed method can have a practical impact on clinical applications by reducing the need for fully-annotated data. As shown in our experiments, it improves the accuracy of medical image segmentation when very few labeled images are available. This could help reduce the workload of radiologists, thereby reducing costs, and provide clinicians with better information for diagnosis. While our empirical evaluation has shown excellent results with very limited data, using fewer annotated images also increases chances of over-fitting outliers in the data (e.g., poor annotations) and produce erroneous or misleading results. A further study on the reliability of medical image segmentation with reduced images is therefore recommended.